# CNN Based Flank Predictor for Quadruped Animal Species

*Vanessa Suessle*[1,*] *Marco Heurich*[2,3,4] *Colleen T. Downs*[5] *Andreas Weinmann*[6] *Elke Hergenroether*[1]

[1] Department of Computer Science, University of Applied Sciences Darmstadt, Schoefferstrasse 3, Darmstadt, Germany
[2] Department of National Park Monitoring and Animal Management, Bavarian Forest National Park, Freyunger Str. 2, 94481 Grafenau, Germany
[3] Department of Wildlife Ecology and Management, Faculty of Environment and Natural Resources, University of Freiburg, Freiburg, Germany
[4] Faculty of Applied Ecology, Agricultural Sciences and Biotechnology, Inland Norway University of Applied Sciences, Evenstad, Norway
[5] School of Life Sciences, University of KwaZulu-Natal, Carbis Road, Scottsville, Pietermaritzburg, South Africa
[6] Department of Mathematics, University of Applied Sciences Darmstadt, Schoefferstrasse 3, Darmstadt, Germany
* E-mail: vanessa.suessle@h-da.de

**Abstract:** The bilateral asymmetry of flanks of animals with visual body marks that uniquely identify an individual, complicates tasks like population estimations. Automatically generated additional information on the visible side of the animal would improve the accuracy for individual identification. In this study we used transfer learning on popular CNN image classification architectures to train a flank predictor that predicts the visible flank of quadruped mammalian species in images. We automatically derived the data labels from existing datasets originally labeled for animal pose estimation. We trained the models in two phases with different degrees of retraining. The developed models were evaluated in different scenarios of different unknown quadruped species in known and unknown environments. As a real-world scenario, we used a dataset of manually labeled Eurasian lynx (*Lynx lynx*) from camera traps in the Bavarian Forest National Park to evaluate the model. The best model, trained on an EfficientNetV2 backbone, achieved an accuracy of 88.70 % for the unknown species lynx in a complex habitat.

## 1  Introduction

In recent years, camera traps have emerged as an important tool for wildlife monitoring in the context of wildlife conservation and management. They are low cost, non-invasive, capture the animal in its natural habitat and can be triggered at any time during night and day. However, when multiple camera traps are deployed, the amount of captured data quickly surpasses what can be manually analyzed. Analyzing such data is time-intensive and prone to errors, especially when conducted manually over several hours [1]. In addition to the common tasks of detection and species identification, researchers can gain additional information from the captured data, such as the visible side of the animal. The additional information can improve main analysis tasks like individual identification, which is of special interest to monitor the population of a particular species in a specific area. It relies on visual characteristics that uniquely identify an individual, for example fur patterns for lynx, zebras (*Equus quagga*) and leopards (*Panthera pardus*) or body marks on manta rays (*Manta alfredi*) and whale sharks (*Rhincodon typus*) [2]. However, a challenge in individual identification arises because the visual characteristics are usually bilaterally asymmetric on the flanks of an animal [3]. Meaning they are independent and both sides of an individual need to be captured for a successful individual identification. If solely each side is captured individually, with no overlap in videos or images, the flanks cannot accurately be assigned to the same individual. To cope with the issue of the bilateral asymmetric flanks, some studies only concentrate on one flank of all individuals for the analysis, not using the full potential of the data [3, 4].

Other studies treat both sides independently as individuals [5], but it bears the risk of overestimating the population, which can bias the results of a monitoring and therewith can negatively affect management decisions. Alternatively statistical methods can be applied for approximations with missing flank footage [6].

Over the past decades, many studies have mentioned the challenge of bilaterally asymmetric individual markings for individual identification [3, 4, 8]. To address this challenge studies on Felidae monitoring often use a dual camera setup from opposing positions [9]. Even though this approach manages to capture both flanks of an animal, the subsequent manual analysis required to sort the captions into the classes on the visible flank remains time-intensive. Therefore, incorporating a component for predicting the visible flank of an animal could save labeling time, prevent overestimating of populations and could help to get the full potential from the captured data. A flank predictor could be embedded in an individual identification pipeline as part of the data preprocessing to increase accuracy by filtering out impossible matches on opposite flanks, reduce computing time and store additional information on the available viewpoints on individuals, because for research on population monitoring not only matching two images as one individual is relevant, but also to clearly identify to images as impossible to match [10]. For existing datasets collected with a single camera at a specific location, it is not possible to retrospectively obtain the information that could have been extracted from a setup with two opposing cameras. Furthermore, future studies with budget limitations and a restricted number of camera traps may have to choose between a larger monitoring area or having dual captions at a camera spot. Apart from the concept of dual camera trap setup, no automatization exists for identifying the visible flank, which could significantly enhance wildlife monitoring in terms of analysis time and accuracy. This study aims to develop a model that can be applied to multiple species and integrated as a modular component in individual identification pipelines. Training a convolutional neural network (CNN) model for flank prediction requires a large amount of labeled data, which is time-consuming to produce manually. In this work an approach is used to automatically derive flank labels from available datasets (Table 1) of quadruped mammals that were labelled with skeletal keypoints (Figure 1). Based on the assumption that even though animals have a high diversity in their appearances, the basic skeletal structure and locomotion of quadrupeds is similar.

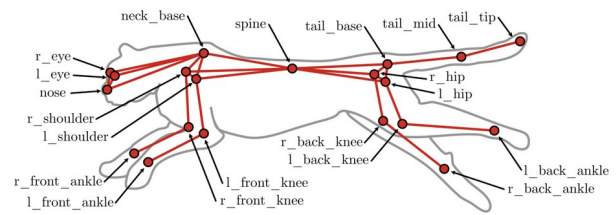

**Fig. 1**: Example labels of body keypoints for pose estimation [7]

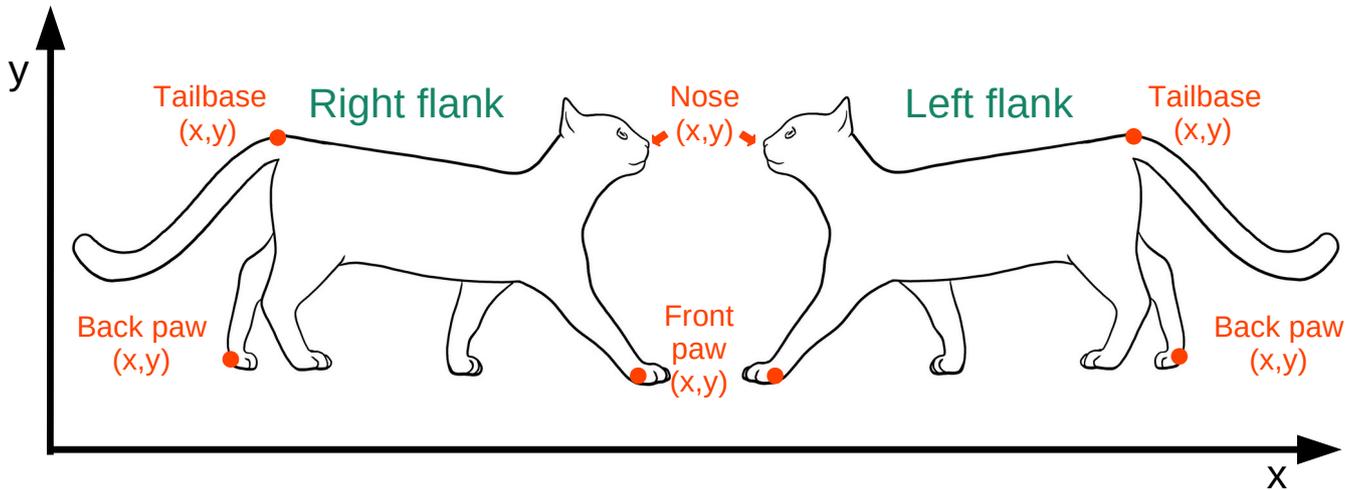

**Fig. 2**: Flank label based on the relative horizontal position of the body keypoints.

## 2   Related Work

Many studies focused on the systematic setup of camera traps to find suitable locations to increase the possibility to capture usable images. Favourable camera spots may be close to water sources, on trails, ridges or locations that were known to be frequently used by the species of interest from former observations or data from collared individuals [3]. The ideal locations highly depend on the topographical aspects and species of interest and should be determined in consultation with local experts [11]. To increase the chances of captions camera spots can be equipped with baits to lure wildlife. A dual setup on each side of the trail increases the chance of capturing both flanks of an individual [3, 8, 9]. For single flank captions it is of interest to define which flank is visible for further analysis. But, even for datasets that were collected in a dual camera trap setup, the manual task to label one of the images of each image pair on their viewpoint, to automatically label the image from the opposite side remains. To benefit the actual analysis objective of the research, which could focus on individual identification or behavioural studies, several studies added interim analysis steps during the data processing that would return additional insight into the data. Those additional tasks that many researches have in common may include the detection [12] and localization of animals in images, counting individuals in herds [13] or pose estimation or tracking on the seen animal [7]. All those approaches are related to the flank prediction focused on in this work and share the challenge of high diversity in the nature of animals. Many of the solutions are bound to a species or a group of species. Pre-trained models mostly exist for domestic or farm animals because of the easier collection of data.

For the task of animal detection a generic model for cross-species application with a high accuracy in detection is the MegaDetector. Its high adaptability to versatile habitats and applications makes it an ideal first component for many analysis pipelines [14]. A similarly versatile model for multi species applications for a generic flank prediction would benefit a multitude of research tasks. Pose estimation is a closely related task to flank prediction. It predicts the pose of an animal shown in a video or image based on predefined body keypoints such as joints or other prominent characteristics like eyes and nose, similar to human pose estimation [15]. Most models are tied to the species the model was trained on. The first model that is applicable to versatile cross-species applications is DeepLab's SuperAnimal-Quadruped [16] model which has been trained on over 40,000 images of four-legged animals. It was planned as an out of the box model to suit many applications without data labelling. A binary classification flank prediction seems less complex than an exact pose estimation on several keypoints, and can be held more generic for cross-species applications. Many pose estimation models rely on video data inputs. Due to higher battery consumption and storage image data is often preferred over video data when collecting data in the wild. Although, meta data from consecutive frames in unlabeled video data can benefit individual identification compared to image data [10].

To the best of our knowledge the only study that tackled the issue of predicting the captured flank of an animal, developed a model that classifies images into different classes combining the species and viewpoint. The class included four different animal species and eight perspectives for zebras and giraffes (*Giraffa reticulata*). The model was trained on combined species-viewpoint classes. Examples of the classes are 'zebra-right', 'zebra-left', 'zebra-front', 'giraffe-left' and 'giraffe-front' with a total of 16 combinations for the species zebra and giraffe. Further treated species in the study were whales and turtles [17]. Combining a species classifier and viewpoint prediction has the downside of limiting the application to the species the model was trained on. The obstacle so far was that training a CNN model requires a large amount of labelled training data, which is not available for the classification of the visible flank of other animal species in images. But several datasets labeled with keypoints on poses for different animal species already exist and could be used to extract labels. The available labeled datasets cover a similar set of body keypoints, including joints and prominent features, and cover the following species: different domestic and farm animals [18], different dog breeds [19], horses (*Equus ferus caballus*) [20], amur tigers (*Panthera tigris tigris*)[5], cheetahs (*Acinonyx jubatus*) [7] and a dataset of 50 different mammalian species [21].

**Table 1**  Datasets with labeled data

| Dataset (Species) | Labeled Annotations | Distribution Left / Right Undefined |
|---|---|---|
| ATRW [5] (Amur tiger) | 2192 | 1166 / 995 / 32 |
| Animal pose dataset [18] (Dog, cat, sheep, cow, horse) | 5660 | 1973 / 2144 / 1543 |
| AP-10K [21] (50 mammalian species) | 8524 | 2730 / 2844 / 2950 |
| Stanford dogs [19] (Dog breeds) | 10,059 | 4482 / 3936 / 1641 |

**Table 2**  Body keypoints sorted into Front and Back class

| Front of Body | Back of Body |
|---|---|
| Ears, nose, head, whiskers, chin, throat, neck, front paws, elbows, shoulder, whithers | Tailbase, tailend, back paws, back knees, hip |

## 3 Methodology

### 3.1 Extraction of Data Labels

We extracted the labels for the data to train the flank predictor model from different available datasets (Table 1) that had labels for the location of predefined body keypoints (Figure 1). The body keypoints from the different datasets varied, but all datasets had labels for prominent features in common on the head, face and tail as well as on the front and back limbs in common that all quadruped mammalian species share. To obtain a large versatile dataset with flank labels to train a generic model that can be applied to different species, we used multiple datasets with its body keypoints from over 50 different animal species and different dog breeds and combined them in one dataset. The total dataset held 26,000 images showing over 31,000 animals. We derived the labels for each image from the body keypoints, based on the assumption that the animals have a quadruped gait, with a mostly walking, not climbing, locomotion. We extracted the flank labels based on the relative position of the relevant body keypoints in the image. The most important body keypoints were the head and tail, but we considered the position of the front and back paws as well, in case the head and tail were not visible. The keypoint classes from the original datasets were split into two groups. One group included the keypoints that were predefined to belong to the front of the animal's body and the other group of keypoints belonged to the back part of the animal. Table 2 shows example bodypoints for each class.

Depending on the position of the x-value of the points we automatically assigned labels in the following way. If the nose and/or front paws of the animal were located farther on the right side of the image than the tailbase and/or back paws, we labeled the image as 'right', meaning that the right flank of the animal in the image is visible. We proceed analogously for the left side (Figure 2). Images for which the front and back class keypoints could not clearly be separated across the x-axis and therefore did not fulfill the condition, were not considered for the further process. Table 1 gives a quantitative overview of labels that were successfully extracted from each of the used dataset. The largest proportion of the dataset consists of dogs. We excluded some species from the AP-10K dataset, because they did not fulfill the requirement of pure quadrupedalism or extensively use climbing for locomotion. For an animal hanging vertically on a tree the label cannot clearly be derived based on the x-value position of body front and body back parts (Figure 9). Furthermore, species that were mostly covered by water were filtered out, because of the invisibility of multiple bodyparts. Excluded species were: hippos (*Hippopotamus amphibius*), otters (*Lutrinae*) and any kind of monkeys (*Simiiformes*).

### 3.2 Training the Model

To develop the flank predictor, we conducted transfer learning on common image classification models. We used ResNet-50 [22], MobileNetV2 [23] and EfficientNetV2-S [24] as backbones, because they are known benchmark models for image classification. All training and validation datasets were preprocessed with the MegaDetector [14] to detect and cut out the annotation of the animal, if bounding boxes on the animal were not already part of the labels. If images contained multiple animals, we processed each annotation individually with an individual flank label. Furthermore, we applied data augmentation techniques to prevent overfitting and to generalize the model. We used random zoom and small random rotations (Figure 5). For the random zoom we either zoomed into the image or out by adding some pixels around the image. For our application case the random rotation was limited to a maximum of 90 degrees, so that the correct flank label and assumptions still apply. The rotation of images strengthens the model's performance and simulates real-world cases, because of movement animals will not always be oriented perfectly in a caption. We did not use the data augmentation technique of flipping images, because the mirroring around an axis would lead to an incorrect label.

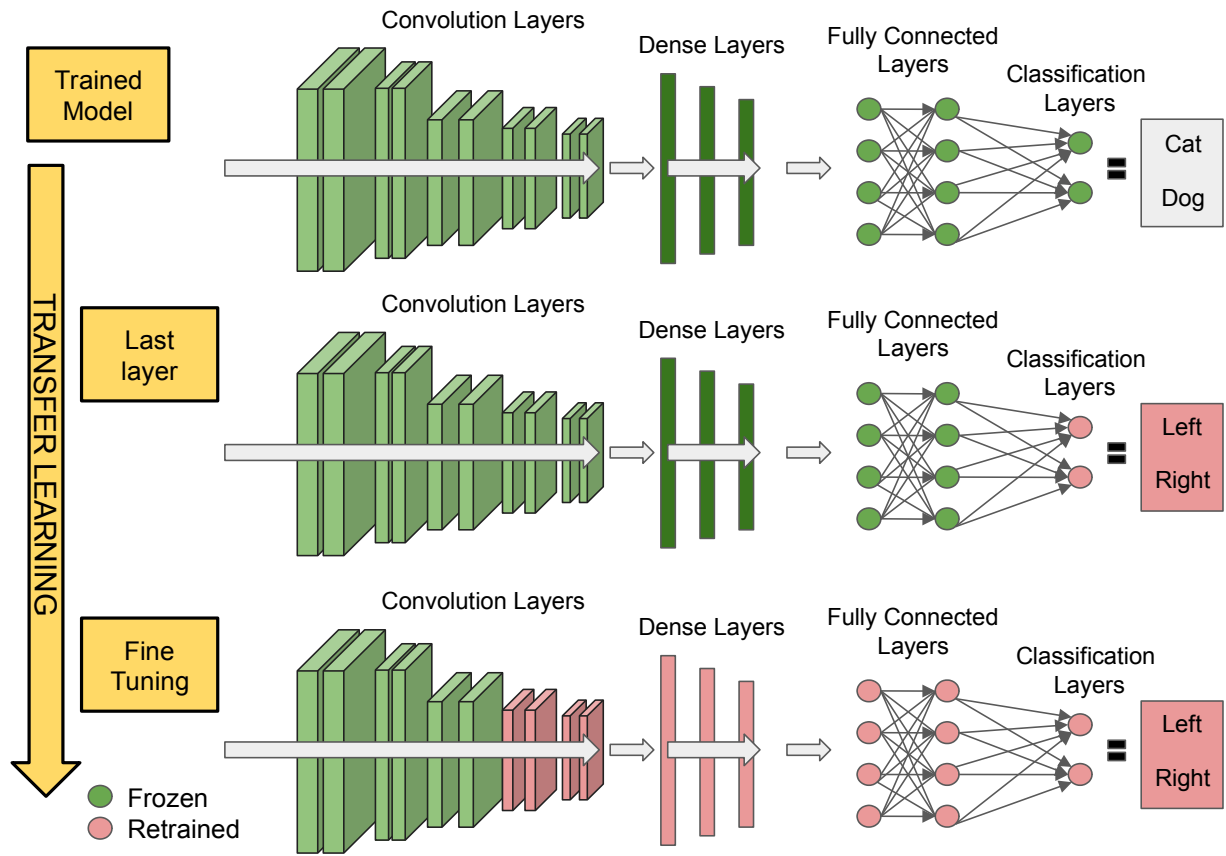

**Fig. 3**: CNN architecture for transfer learning.
*Top:* Original architecture.
*Middle:* All layers frozen, but last layer.
*Bottom:* Only the first layers are frozen and the later layers are retrained.

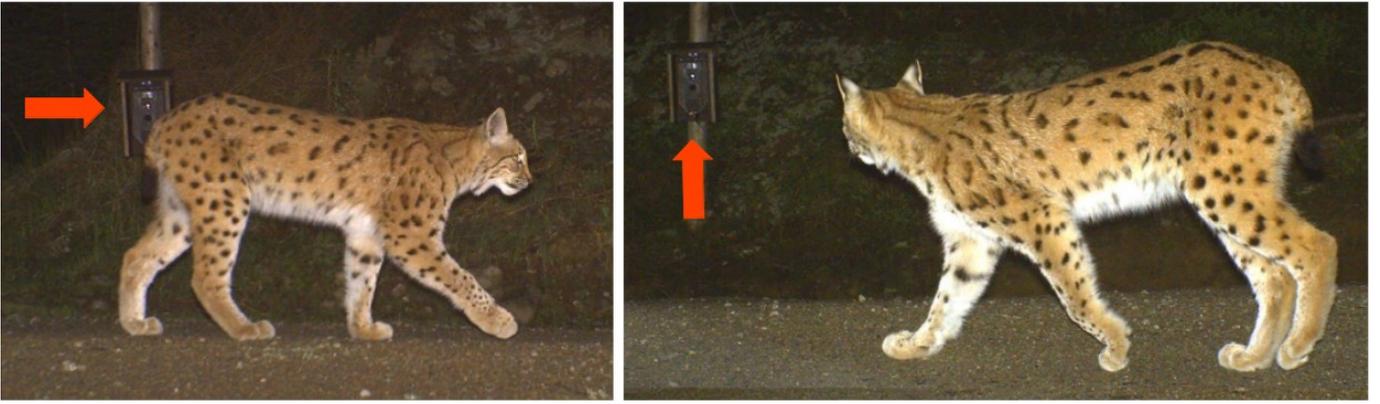

**Fig. 4**: Dual camera trap setup to capture both flanks of an individual of a lynx.

The idea of transfer learning is to use the weights and layers of existing models and retrain the model on newly labeled data. During this process different layers of the existing models can be frozen, meaning that the parameters in those layers will not be updated. The first layers of a model extracts basic features from the processed images. The layers in the rearmost of the network are more specialized than the ones in the front and learn more complex features. The last fully connected layer fulfills the task of classification. It returns a "probability" for each of the classes of the processed image belonging to that class, from which the maximum is assigned as the predicted class. In this study the last layer is a binary classification layer, sorting into the classes 'right' and 'left' stating which flank of the animal is visible in the image. This study combined two approaches in two training phases. The first training phase kept most of the models' layers untouched and only added the last layer for the binary classification. The second phase of training allowed more layers to be retrained by freezing the first half of the layers of each model and adding the fully connected binary classification layer as the last layer. The schematic outline of the two training phases is illustrated in Figure 3.

During the first training phase all layers, but the last layer were frozen. In this phase we trained the models in 15 epochs with a batch size of 32. We added a fully connected dense layer as a last layer to the model backbones. The last binary classification layer was trained on the new labeled data and returns the probability of the processed image belonging to each of the two classes 'left' and 'right'. In the second training step we fine-tuned the model and the weights of multiple later layers in the model were unfrozen and retrained. During retraining, the last layers are fine-tuned and learn features that are more related to the dataset at hand, compared to the earlier learned more generic features. During the fine-tuning phase the first half of layers of the backbones remain untouched to retain their ability to extract basic features. In the second training phase we trained models for further 15 epochs with a batch size of 32 as well. Figure 7 shows the training process over the epochs. The step between the first and second training phase is marked with a red line.

## 4 Case Study

We evaluated the model in two different scenarios on unknown datasets. The two different scenarios covered different degrees of complexity in terms of lighting, quality and poses of the animals. The species present in the two validation dataset were excluded from the training set. For the species unknown to the model we selected Felidae for this research, because many studies focus on the individual identification of Felidae due to declining populations [25].

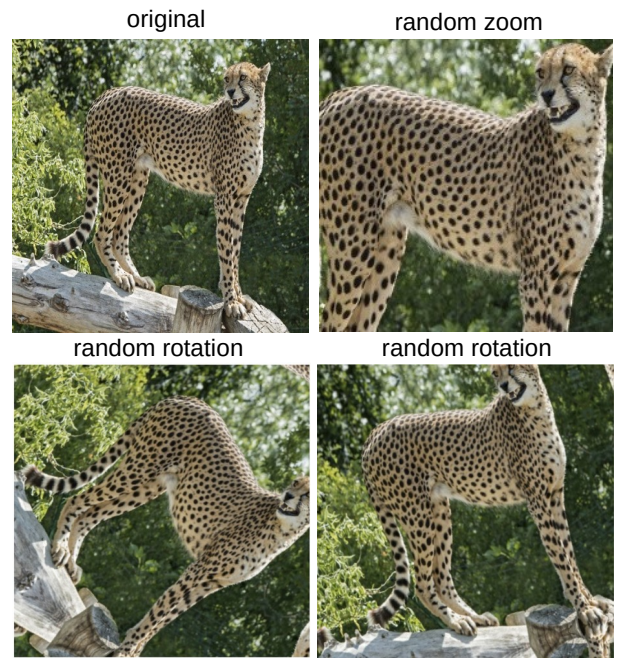

**Fig. 5**: Data augmentation on the training dataset. The annotations of the cheetah were randomly zoomed in and randomly rotated by less than 90 degrees.[21]

### 4.1 Leopard and Bobcat Dataset

The first validation dataset consists of leopards and bobcats (*Lynx rufus*) and the second datasets covers Eurasian lynx (Figure 6). We split the leopard and bobcat validation dataset from the AP-10K dataset [21], which covers over 50 species. The animals are centered in the images with good viewpoints. The species leopard and bobcat were exclusive in the validation set and not presented to the model during the training process. The leopard and bobcat validation datasets covered similar conditions in terms of quality and lighting as the training dataset.

### 4.2 Lynx Dataset from Cameratraps

The second dataset was collected in the Bavarian Forest National Park and filtered for lynx with two camera traps positioned on opposite sides of a trail to capture individuals from both sides (Figure 4) [26]. The dataset were manually labeled for the viewpoint on the animal and preprocessed with the MegaDetector to detect the animal returning a bounding box. The dataset consists of 4134 images with 2045 being in the class left flank and 2089 in right flank.

4Camera Traps, AI, and Ecology
© Copyright resides with the authors

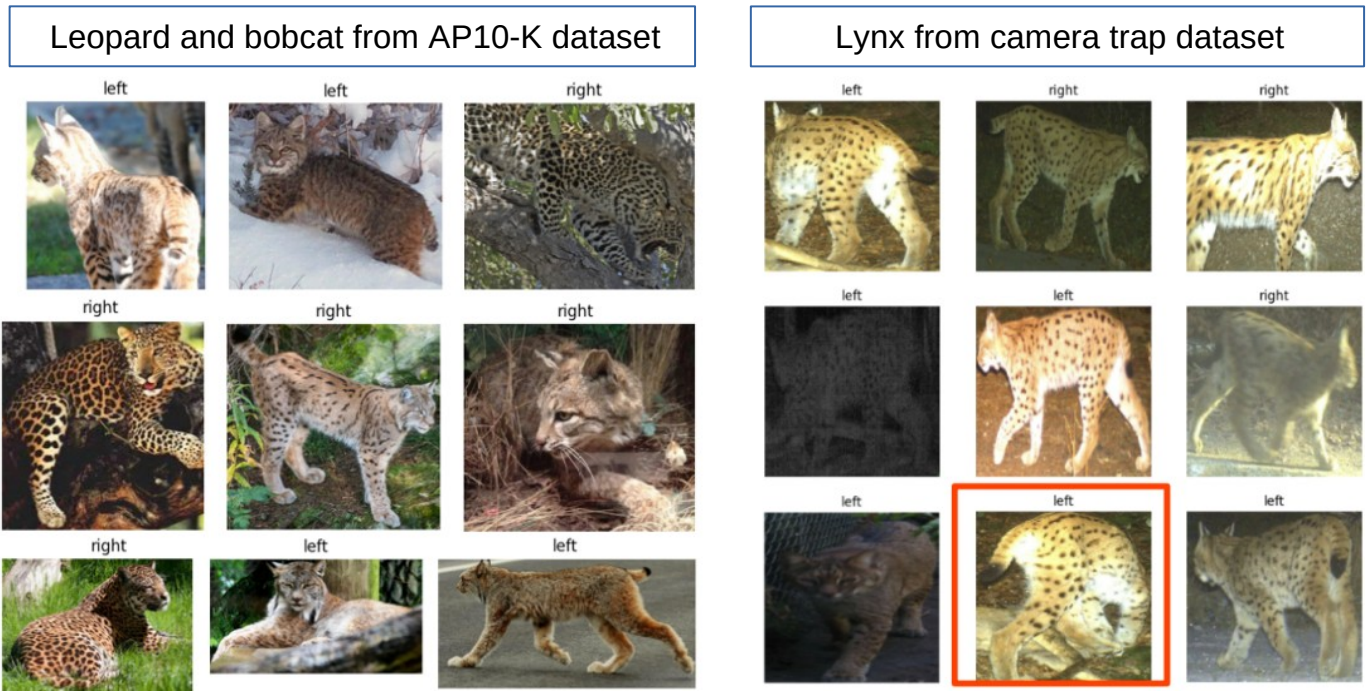

**Fig. 6**: Predicted flank labels on the valdation datasets for the unseen species: lynx, bobcat and leopard. The prediction highlighted in red is incorrect. Image reference: Bobcat and leopard: [21]; Eurasian lynx: Bavarian Forest National Park

**Table 3** Accuracies obtained by the models on the datasets

| Model | Total number / convolutional layers | Frozen layers | Train Accuracy | Validation Accuracy On Unseen Species Leopard / Bobcat | On Unseen Species **Lynx** |
|---|---|---|---|---|---|
| | | | Before fine tuning (one trainable layer) | | |
| MobileNetV2 [23] | 35 | 1 | 53.29 % | 48.87 % | **63.09 %** |
| ResNet-50 [22] | 50 | 1 | 56.77 % | 57.89 % | **39.01 %** |
| EfficientNetV2-S [24] | 42 | 1 | 52.72 % | 44.36 % | **44.65 %** |
| | | | **Final models** after fine-tuning (half of model's layers frozen) | | |
| MobileNetV2 | 35 | 18 | 90.02 % | 97.74 % | **71.63 %** |
| ResNet-50 | 50 | 23 | 92.79 % | 97.74 % | **77.59 %** |
| EfficientNetV2-S | 42 | 20 | 96.34 % | 98.50 % | **88.70 %** |

## 5 Results

We retrained and fine-tuned three models based on the ResNet-50, MobileNetV2 and EfficientNetV2-S backbones using the derived flank labels from datasets including different mammalian species. The models output a prediction on a binary classification task on the visible flank. Exemplary presentations of predictions on images from the lynx dataset and the leopard and bobcat dataset are shown in Figure 6.

### 5.1 Data Augmentation

The graph in Figure 7 shows the training process exemplary for the ResNet-50 with and without data augmentation validated on the lynx dataset. The data augmentation increased the performance of the ResNet-50. Although, the training accuracy without data augmentation was higher than for the model trained with augmented images, the model trained with augmentation performed better on the validation dataset, preventing the model from overfitting.

### 5.2 Fine-Tuning

The red vertical bar marks the step between the different training phases for which different amounts of layers were frozen. The fine tuning of the models significantly increased the performance of the models. The graph in Figure 8 shows an analysis on how the number of frozen layers during fine-tuning affects the performance. The number of layers was chosen to fit the residual blocks of the network architectures. When freezing more than half of the network's layer the performance stagnated. For the further analysis we froze the first half and retrained the second half of each model's layers in the fine-tuning process.

### 5.3 Performance in Case Study

Table 3 summarises the different models validated on the different validation datasets.

The fine-tuned models generally performed better than the basic models for which solely the rearmost layer of the network was retrained (Figure 3). On the lynx dataset with different light conditions, habitats and camera setups, the basic model predictions were poor with accuracies of lower than 50 %. The basic models performed better on the leopard and bobcat validation set, originating from the AP-10K dataset. Still, the predictions are too unreliable for real-world applications for camera trap analysis. The fine-tuning training phase significantly improved the performance of the models for unseen species (Figure 7). The accuracy doubled for most of the backbone models. The best performing model on the task of flank prediction for the previously unseen lynx species in unknown habitat is the EfficientNetV2-S. It reached an accuracy of 88.70% compared to its competitors MobileNetV and ResNet-50 that reached 71.63% and 77.59% (Table 3).

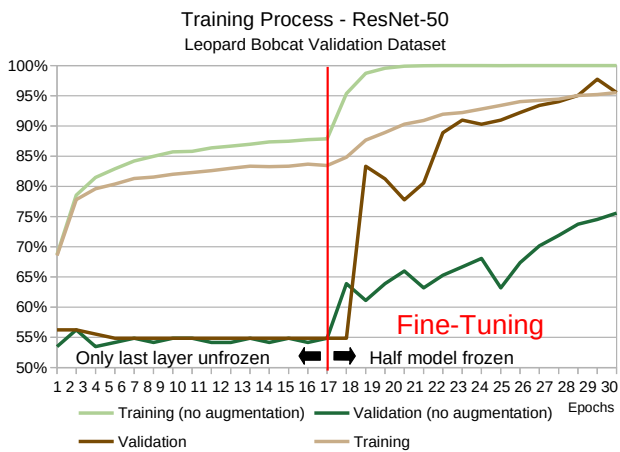

**Fig. 7**: Performance of the model during the training process. The red line marks the switch from first to the second training phase for which less layers were frozen during transfer learning.

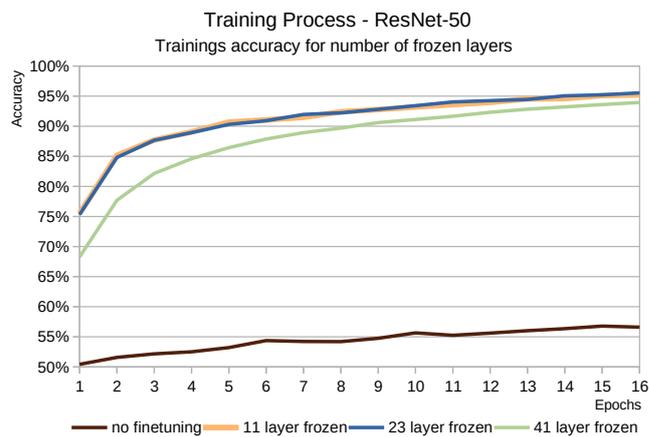

**Fig. 8**: During the training process different amounts of layers of the ResNet-50 were frozen. The performance stagnated for more than half of the layers being frozen.

All models performed better on the validation set with bobcats and leopards compared to the dataset of lynx. From the 135 images all models classified over 130 images correctly.

We assume that the characteristics of the dataset can explain the generally better performance for the bobcat and leopards compared to the lynx. The bobcat and leopard validation dataset was not included in the training phase, but the data originated from the AP-10K dataset [21] that was used during training. The images therefore resemble one another in terms of lighting, quality and environments, while the lynx dataset includes images captured in a different environment and included images during nighttime.

Images with snow in the background or images that include foreign objects like camera traps and collars tend to be classified incorrectly (Figure 10). If prey is present the flank prediction in some cases was done for the prey body. Another cause for incorrect predictions are animals in twisted positions (prediction highlighted with red box in Figure 6). The reason for the incorrect classification lies in the labeled data. The automatically extracted data labels do not cover all eventualities. In the above shown case the head and face are not visible and the front and hind paws are close together, for which the body keypoints from the classes for the front and the back of the animal cannot clearly be separated. We excluded such images in the label extraction process and therewith from the training dataset. Animals in such poses could not be properly learned by the models.

## 6 Conclusions

In this study we trained and compared flank prediction models for quadruped species using transfer learning on different backbone network architectures (Figure 3). The data labels were derived from datasets labeled on body keypoints for different mammalian species (Figure 2). To further improve the models, pose or flank labeled data during nighttime is required to make the models more robust in low light conditions. Data from a wider variety of habitats could make the model more robust towards new application cases.

The automatic label extraction was limited to the predefined conditions of the keypoints relative position to each other and their visibility in the image. Most cases were covered by one of the conditions, but some images of animals in twisted poses could not clearly be sorted into one of the classes of the visible flank. Examples include that one of the flanks was better visible than the other, but the position of the head and tail implied the opposite label or images were the animal is in unusual positions that do not meet the conditions made for an automatized extraction. The example presents a tiger vertically climbing a tree-like object. In this position the assumption of a quadruped gait is not met (Figure 9).

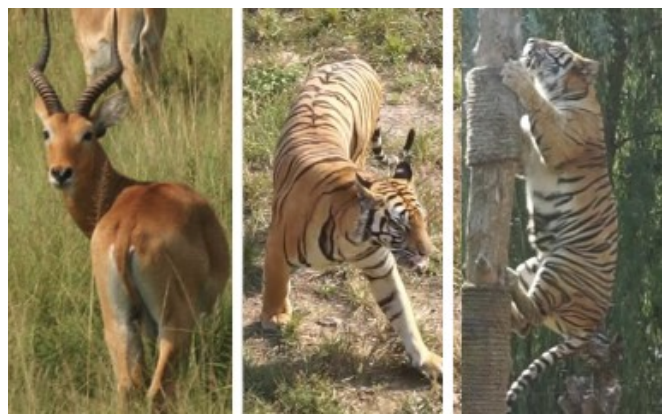

**Fig. 9**: Animals in twisted poses for which the conditions were not fulfilled or led to a wrong autolabelling of body keypoints.

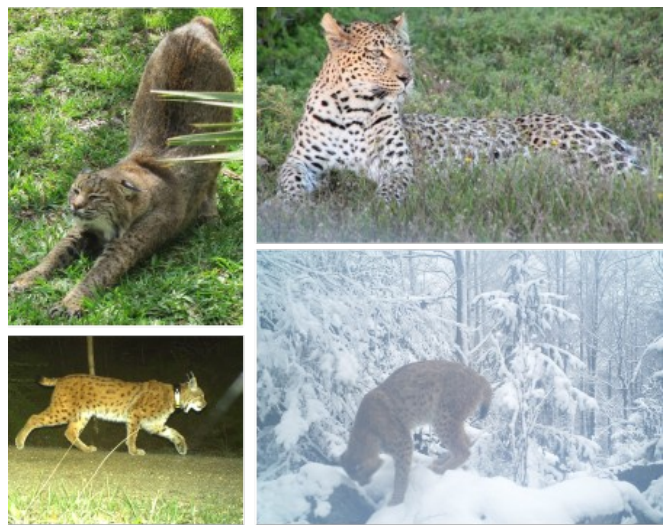

**Fig. 10**: Misclassified images from the leopard and bobcat AP-10K dataset [21] and the lynx camera trap dataset.

For future approaches the dataset could be extended with manually labeled data for twisted and unusual positions. Nevertheless, with animals being non rigid objects in versatile poses and habitats, predefined conditions or manually labeled data will not exhaustively cover all possibilities of animal poses for different species.